%% file: main.tex
\documentclass[nohyperref]{article}

     \PassOptionsToPackage{numbers, compress}{natbib}

\usepackage{microtype}
\usepackage{graphicx}
\usepackage{subfigure}
\usepackage{natbib}
\usepackage{booktabs} % for professional tables

\usepackage{hyperref}

\usepackage[accepted]{icml2021}

\usepackage{times}
\usepackage{latexsym}

\usepackage[utf8]{inputenc} % allow utf-8 input
\usepackage[T1]{fontenc}    % use 8-bit T1 fonts
\usepackage{hyperref}       % hyperlinks
\usepackage{url}            % simple URL typesetting
\usepackage{booktabs}       % professional-quality tables
\usepackage{amsfonts}       % blackboard math symbols
\usepackage{nicefrac}       % compact symbols for 1/2, etc.
\usepackage{microtype}      % microtypography
\usepackage{amsmath,amssymb}
\usepackage{epsfig,graphicx,subfigure,caption}
\usepackage{multirow,booktabs, hhline}
\usepackage[algo2e]{algorithm2e}
\usepackage{amsmath, bm}
\usepackage{wrapfig}
\usepackage{lipsum}
\usepackage{caption}
\usepackage{footnote}
\usepackage[bottom]{footmisc}
\usepackage[toc,page]{appendix}

\usepackage[textsize=tiny]{todonotes}

\icmltitlerunning{VLUE: A Multi-Task Benchmark for Evaluating VIsion-Language Pre-training}

\newcommand{\specialcelll}[2][l]{%
  \begin{tabular}[#1]{@{}l@{}}#2\end{tabular}}

\begin{document}

\twocolumn[
%\icmltitle{VLUE: A Multi-Task Multi-Dimension Benchmark for Evaluating \\ Performance, Generalization, and Efficiency of Vision-Language Pre-training }

\icmltitle{VLUE: A Multi-Task Benchmark for Evaluating  Vision-Language Models}

%VLUE: A Benchmark for Evaluating Vision-LanguagePre-training, Understanding, and Generalization


\icmlsetsymbol{equal}{*}

                \begin{icmlauthorlist}
                \icmlauthor{Wangchunshu Zhou}{equal,bytedance}
                \icmlauthor{Yan Zeng}{equal,bytedance}
                \icmlauthor{Shizhe Diao}{equal,sch}
                \icmlauthor{Xinsong Zhang}{equal,bytedance}
                \end{icmlauthorlist}
                
                \icmlaffiliation{bytedance}{ByteDance AI Lab}
                %\icmlaffiliation{comp}{Company Name, Location, Country}
                \icmlaffiliation{sch}{The Hong Kong University of Science and Technology}
                
                \icmlcorrespondingauthor{Wangchunshu Zhou}{wcszhou@outlook.com}
                %\icmlcorrespondingauthor{Firstname2 Lastname2}{first2.last2@www.uk}

\icmlkeywords{Machine Learning, ICML}

\vskip 0.3in
]



\printAffiliationsAndNotice{\icmlEqualContribution} % otherwise use the standard text.

\newcommand{\benchmark}{VLUE\xspace}

\input{sections/0_abstract}
\input{sections/1_introduction}
\input{sections/2_related}
\input{sections/3_benchmark}
\input{sections/4_experiments}
\input{sections/5_conclusion}

\bibliography{new}
%\bibliographystyle{unsrtnat}

%%%%%%%%%%%%%%%%%%%%%%%%%%%%%%%%%%%%%%%%%%%%%%%%%%%%%%%%%%%%%%%%%%%%%%%%%%%%%%%
%%%%%%%%%%%%%%%%%%%%%%%%%%%%%%%%%%%%%%%%%%%%%%%%%%%%%%%%%%%%%%%%%%%%%%%%%%%%%%%
% APPENDIX
%%%%%%%%%%%%%%%%%%%%%%%%%%%%%%%%%%%%%%%%%%%%%%%%%%%%%%%%%%%%%%%%%%%%%%%%%%%%%%%
%%%%%%%%%%%%%%%%%%%%%%%%%%%%%%%%%%%%%%%%%%%%%%%%%%%%%%%%%%%%%%%%%%%%%%%%%%%%%%%
\newpage
\appendix
\onecolumn
\section{Actual Inference Time}

We present the actual inference time of compared models in Table 3.
%%%%%%%%%%%%%%%%%%%%%%%%%%%%%%%%%%%%%%%%%%%%%%%%%%%%%%%%%%%%%%%%%%%%%%%%%%%%%%%
%%%%%%%%%%%%%%%%%%%%%%%%%%%%%%%%%%%%%%%%%%%%%%%%%%%%%%%%%%%%%%%%%%%%%%%%%%%%%%%

\begin{table*}[ht]
\centering
\begin{tabular}{lccccccc}
\hline
\textsc{Model} & \textsc{ViLBERT} & \textsc{LXBERT} & \textsc{UNITER} & \textsc{VL-T5}  & \textsc{ALBEF} & \textsc{X-VLM} & \textsc{METER} \\
\midrule
\textsc{Performance}  & 57.46 & 55.71  & 57.20 & 60.08 & 62.35 & 62.86 & 63.62 \\
\textsc{\begin{tabular}[c]{@{}c@{}}Infer. Time \\ (O.D. Time)\end{tabular}}  & 
\begin{tabular}[c]{@{}c@{}}152.16 \\ (121.89)\end{tabular} &
\begin{tabular}[c]{@{}c@{}}154.36 \\ (121.89)\end{tabular} & \begin{tabular}[c]{@{}c@{}}157.39 \\ (121.89)\end{tabular} &  
\begin{tabular}[c]{@{}c@{}}164.51 \\ (121.89)\end{tabular} & 63.50 & 79.50 & 64.80 \\
\hline
\end{tabular}
\caption{The performance and inference time across seven models.
\textsc{Infer.Time} and \textsc{O.D. Time} refer to the inference and object detection time cost in terms of millisecond, respectively.
}
\label{tab:efficiency-performance-tradeoff-figures}
\end{table*}

\end{document}

%% file: sections/0_abstract.tex
\begin{abstract}
Recent advances in vision-language pre-training (VLP) have demonstrated impressive performance in a range of vision-language (VL) tasks. 
However, there exist several challenges for measuring the community's progress in building general multi-modal intelligence. 
First, most of the downstream VL datasets are annotated using raw images that are already seen during pre-training, which may result in an overestimation of current VLP models' generalization ability. 
Second, recent VLP work mainly focuses on absolute performance but overlooks the efficiency-performance trade-off, which is also an important indicator for measuring progress.

To this end, we introduce the Vision-Language Understanding Evaluation (\benchmark) benchmark, a multi-task multi-dimension benchmark for evaluating the generalization capabilities and the efficiency-performance trade-off  (``Pareto SOTA'') of VLP models.
We demonstrate that there is a sizable generalization gap for all VLP models when testing on out-of-distribution test sets annotated on images from a more diverse distribution that spreads across cultures.
Moreover, we find that measuring the efficiency-performance trade-off of VLP models leads to complementary insights for several design choices of VLP.
We release the \benchmark benchmark\footnote{The benchmark is publicly available at \url{https://vlue-benchmark.github.io}. The data and codes used for training baseline models are available at \url{https://github.com/MichaelZhouwang/VLUE}.} to promote research on building vision-language models that generalize well to more diverse images and concepts unseen during pre-training, and are practical in terms of efficiency-performance trade-off.

\end{abstract}

%%from the same distribution or even overlapping with image-text pairs used for VL pre-training. 
%Second, while comparing absolute performance of models is essential, the efficiency-performance trade-off is also an important indicator for measuring the progress but under-explored in the vision-language community.

%We demonstrate that while VL pre-trained models achieves great performance on VL tasks annotated on in-distribution images, there is still a sizable generalization gap when testing on images from a more diverse distribution that spreads across cultures. 
%Moreover, we find that in addition to comparing the absolute performance, measuring the efficiency-performance trade-off of VL models leads to complementary insights for several design choices of pre-training vision-language models. 

%compared to that in the original in-distribution test sets. 
% ``Pareto SOTA'' 
%Finally, the evaluation protocols of recent VL models are inconsistent, making it complicated to compare among state-of-the-art results. 

%% file: sections/1_introduction.tex
\section{Introduction}
Building systems that can demonstrate their visual understanding by generating or responding to natural language in the context of images has been a long-standing goal in the field of artificial intelligence and cognitive science~\citep{margaret2008mind}. 
The approaches and corresponding tasks have come to be referred to under the common banner of ‘vision-and-language’~\citep{DBLP:conf/nips/LuBPL19}.
Vision-Language Pre-training (VLP), the paradigm of pre-training on large-scale parallel image-text pairs and then fine-tuning on vision-language (VL) tasks, has achieved state-of-the-art performance on a wide range of VL tasks. 
It has transformed the landscape of vision-and-language research and is considered to be a critical step for building general multi-modal intelligence. 
Over the last two years, a great number of research studies ~\citep{DBLP:conf/nips/LuBPL19,DBLP:conf/emnlp/TanB19,chen2020uniter,DBLP:conf/aaai/LiDFGJ20,DBLP:journals/corr/abs-1908-03557,DBLP:conf/eccv/Li0LZHZWH0WCG20, DBLP:conf/icml/ChoLTB21,DBLP:conf/cvpr/ZhangLHY0WCG21,li2021align,DBLP:conf/icml/JiaYXCPPLSLD21,DBLP:journals/corr/abs-2111-08276,DBLP:journals/corr/abs-2111-02358,DBLP:journals/corr/abs-2108-10904} have been conducted in the field of vision-language pre-training.
Nevertheless, it becomes increasingly complicated to track the \textit{real} progress that the vision-language community is actually making because of two major problems in the common practice for evaluating and reporting new studies in the field. We elaborate them as follows.

First, most datasets of downstream VL tasks used for evaluating VLP models are annotated with images collected from the COCO~\citep{DBLP:conf/eccv/LinMBHPRDZ14} or Visual Genome (VG)~\citep{DBLP:journals/ijcv/KrishnaZGJHKCKL17} dataset\footnote{A few studies exclude the images used in some downstream datasets from their pre-training data. However, this does not fully resolve this issue because some datasets are not annotated within a specific subset. Also, even the images are not seen during pre-training, they are still from the same distribution.}.
The problem is that most (if not all) VLP models are pre-trained on image-text pairs from COCO and VG datasets. Therefore, VLP models have already seen the images in the downstream datasets and their captions before fine-tuning and evaluating on them. As such, fine-tuning and evaluating VLP models on these datasets only measures their transferability in the setting where at least a part of data distribution (i.e., image distribution) remains the same while the label distribution shifts. This is a special case of transfer learning and is unlikely to be met in practice.
Consequently, evaluating on these ``in-domain'' datasets will lead to a biased and probably overestimated transfer and generalization ability of VLP models.

Second, most recent studies on VLP mainly focus on the absolute performance improvement but ignore the efficiency-performance trade-off, which is also very important for the application of VLP models in real-world applications. 
This issue is likely to be more severe because recently supersized VLP models are pushing the state-of-the-art (SOTA) of many VL tasks to a new level, making it impossible for most researchers with moderate computation resources to reach results exceed or comparable with SOTA. This phenomenon is common in the field of natural language processing (NLP) and most studies are instead pursuing improvement on other dimensions such as the efficiency-performance trade-off~\citep{DBLP:journals/corr/abs-2110-07038,DBLP:journals/corr/abs-2111-05193,DBLP:conf/emnlp/JiaoYSJCL0L20,DBLP:conf/emnlp/XuZGWZ20,DBLP:journals/corr/abs-2202-07101,DBLP:journals/corr/abs-2202-07105,DBLP:journals/corr/abs-2106-04570,DBLP:conf/emnlp/XuZG0MW21,xu2021beyond}. We refer to the goal of these studies as ``Pareto SOTA'' following~\citep{DBLP:journals/corr/abs-2110-07038}, which means that there is no other model currently better than it on all the dimensions of interest such as performance and efficiency. Therefore, we believe it is necessary to measure and report performance-efficiency trade-off of VLP models to promote and facilitate research in the field.

In addition, the evaluation protocol used in recent VLP studies are not consistent enough and different studies report results on a different set of tasks, datasets, or experimental settings. Consequently, it is complicated for researchers to compare their methods to existing ones and for the VLP community to track progress. This is because the lack of a standard evaluation benchmark like GLUE~\citep{wang2018glue} and SuperGLUE~\citep{wang2019superglue} for natural language understanding research and XTREME~\citep{hu2020xtreme} and XGLUE~\citep{liang-etal-2020-xglue} for multi-lingual generalization of pre-trained models. 

To address these problems and promote research on truly generalizable and practical VLP, we introduce the \textbf{V}ision-\textbf{L}anguage \textbf{U}nderstanding \textbf{E}valuation (\textbf{\benchmark}) benchmark. \benchmark is the first multi-task benchmark focusing on vision-language understanding that covers a set of fundamental VL tasks including image-text retrieval, visual question answering, visual reasoning, and visual grounding, and maintains a leaderboard tracking the performance of representative studies and new methods on VLP. More importantly, \benchmark includes a newly annotated private out-of-distribution (OOD) test set for each representative VL task. In contrast to standard datasets for these tasks that are annotated on COCO/VG images, our private OOD test sets are annotated on images from the MaRVL~\citep{DBLP:conf/emnlp/0001BPRCE21} dataset where images are manually collected across cultures by native speakers from different countries. This ensures that the image distribution in our OOD test sets differs from that of COCO/VG images. Moreover, we carefully control the annotation protocol for our OOD test sets to be identical to the original in-domain datasets. As such, the label distribution in our OOD test sets is roughly the same as the original test set but the image distribution differs. This enables us to better measure the \textit{true} generalization and transferability of VLP models. In addition, we also encourage researchers to measure and compare the efficiency-performance trade-off when reporting new studies in the field of VLP. To facilitate that, we measure the efficiency-performance trade-off of representative VLP models in \benchmark to track a Pareto SOTA landscape for VLP research. 
In general, in contrast to conventional benchmarks that only capture the single performance metric, \benchmark is a multi-dimension benchmark that takes multiple dimensions including performance, generalization ability, and efficiency into account. Hopefully, this will promote research on VLP models that are environmentally friendly and practical for real-world applications.

We evaluate a range of representative VLP models on \benchmark to facilitate future research and analyze their generalization ability and efficiency-performance trade-off with respect to several key design choices. We find that there is a sizable generalization gap for all VLP models when evaluating on new examples annotated with images from in-the-wild distribution. Also, compared to focusing on a single dimension (i.e., absolute performance), measuring the generalization ability of different models can lead to complementary and even controversial conclusions. We also find that models with similar performance may result in completely different positions in the Pareto front measuring the efficiency-performance trade-off of VLP models, which also demonstrates the necessity of a multi-dimension benchmark for evaluating VLP models. 

In sum, our contributions are the following: (i) We release a vision-language benchmark consisting of 4 representative VL tasks, each equipped with a private test set annotated on images from wild distribution; (ii) We provide an online platform and leaderboard for the evaluation and comparison of VLP models and provide a set of strong baselines, which we evaluate across all tasks, and release code to facilitate adoption; (iii) We evaluate the efficiency-performance trade-off of representative VLP models and build a Pareto SOTA landscape for current VLP research; (iv) We provide an extensive analysis of the generalization ability and the efficiency-performance trade-off of representative VLP models.

%% file: sections/2_related.tex
\section{Related Work}
%%%% zengyan: 看到点语法错误, 直接改了

The \textit{pre-training then fine-tuning} paradigm~\citep{peters2018elmo,radford2018improving,devlin2018bert,liu2019roberta,radford2019language,raffel2019exploring,lewis2019bart,brown2020language,dong2019unified,zhou2021improving,zhou2020pre,DBLP:conf/acl/FuZX0022} has substantially advanced the state-of-the-art of natural language processing on a wide range of NLP tasks~\citep{wang2018glue,wang2019superglue,warstadt2018neural,socher2013recursive,dolan2005automatically,agirre2007semantic,williams2018broad,rajpurkar2016squad,dagan2006pascal,lin2019commongen,xu2020matinf,xu2021blow}. 
The success motivates researchers to adopt this paradigm to improve vision-language tasks. 
The idea of vision-language pre-training is to pre-train a general-purpose vision-language model on large-scale parallel image-text pairs and then fine-tune on downstream vision-language tasks.

\begin{table*}[ht]
\centering
\resizebox{\textwidth}{!}{%
\begin{tabular}{l l l r r r r l}
\toprule
Task & Dataset & Image Domain & $|$Train$|$ & $|$Dev$|$ & $|$Test$|$ & $|$OOD Test$|$ & Metric  \\ \midrule
% 这里我给的是 #images
Image-Text Retrieval & MSCOCO & COCO & 566,747  & 25,010 & 25,010 & 27,796 & R@1 \\
Image Captioning & MSCOCO  & COCO & 566,747  & 25,010 & 25,010 & 27,796 & BLEU/CIDER \\
Visual Grounding & RefCOCO+  & COCO & 120,191  & 10,758 & 10,615 & 1,313 & Accuracy \\
Visual Reasoning & NLVR2 & Google Images\footnote{with queries from ImageNet synsets} & 86,373 & 6,982 & 6,967 & 5,662 & Accuracy \\
Visual Question Answering & VQA 2.0 & COCO & 443,757 & 214,354 & 447,793 & 11,942  & Accuracy \\

\bottomrule
\end{tabular}
}
\caption{Characteristics of the datasets in \benchmark.}
\label{tab:tasks}
\end{table*}

The approaches of vision-language pre-training can be categorized according to two major criteria. 
The first dimension is how the visual input is represented in the vision-language model. They are typically two approaches.  
Most existing methods~\cite{DBLP:conf/emnlp/TanB19, DBLP:conf/nips/LuBPL19, DBLP:journals/corr/abs-1908-03557, DBLP:conf/aaai/LiDFGJ20, chen2020uniter, DBLP:conf/eccv/Li0LZHZWH0WCG20, gan2020large, li2020unimo} represent an image by dozens of object-centric features of regions of interest which are identified and extracted by object detection. They either utilize pre-trained object detectors~\cite{ren2015faster, anderson2018bottom} or conduct object detection on-the-fly in the pre-training process~\cite{su2019vl, xu2021e2e}. More recently, a number of work~\citep{kamath2021mdetr,yang2021crossing} explored integrating detection into an end-to-end pre-training procedure. 
The other approaches take raw image pixels as vision input, and extract overall image features with convolutional network~\cite{huang2020pixel, huang2021seeing} or vision transformer~\cite{kim2021vilt, li2021align}.

The second dimension is how the vision and language modalities interact with each other in the vision language model, which leads to two categories: early fusion models and late fusion models. 
Late fusion models such as CLIP~\citep{DBLP:conf/icml/RadfordKHRGASAM21} and  ALIGN~\citep{DBLP:conf/icml/JiaYXCPPLSLD21} encode images and texts separately with a dual encoder architecture, and use cosine similarity or a linear projection layer to model the cross-modality interaction. The dual encoder design is effective for retrieval tasks. 
However, late fusion makes the interaction cross modalities too simple to handle tasks that require complex reasoning, such as visual reasoning and visual question answering. The early fusion models~\citep{DBLP:conf/nips/LuBPL19,DBLP:conf/emnlp/TanB19,DBLP:journals/corr/abs-1908-03557,DBLP:conf/icml/ChoLTB21,DBLP:conf/cvpr/ZhangLHY0WCG21,li2021align,DBLP:journals/corr/abs-2111-08276,DBLP:journals/corr/abs-2111-02358,DBLP:journals/corr/abs-2108-10904} instead use a deep fusion encoder with cross-modal attention to improve cross-modal interaction, leading to improved performance for vision-language understanding tasks.

In addition,~\citet{DBLP:conf/eccv/CaoGCY0020} developed VALUE (short for vision-and-language understanding evaluation), a suite of probing tasks aiming to understand the inner workings of VLP models. The tasks included in \benchmark are instead realistic vision and language tasks and have real-world applications. Therefore, \benchmark can better serve as a standard benchmark for the evaluation of VLP models. Moreover,~\citet{li2021value} developed another VALUE benchmark where the ``V'' stands for video. It focuses on video-and-language tasks whereas \benchmark focuses on image-and-language tasks. Recently,~\citet{su2021gem} introduced the GEM benchmark. GEM is a multimodal benchmark that focuses on both image-language tasks and video-language tasks. Different from \benchmark, GEM focuses on multilingual multimodal models and only considers the image-text retrieval task and image captioning task.

%% file: sections/3_benchmark.tex
\section{\benchmark}

\begin{figure*}[ht]
%\vskip -0.2in
\begin{center}
\centerline{\includegraphics[width=0.8\textwidth]{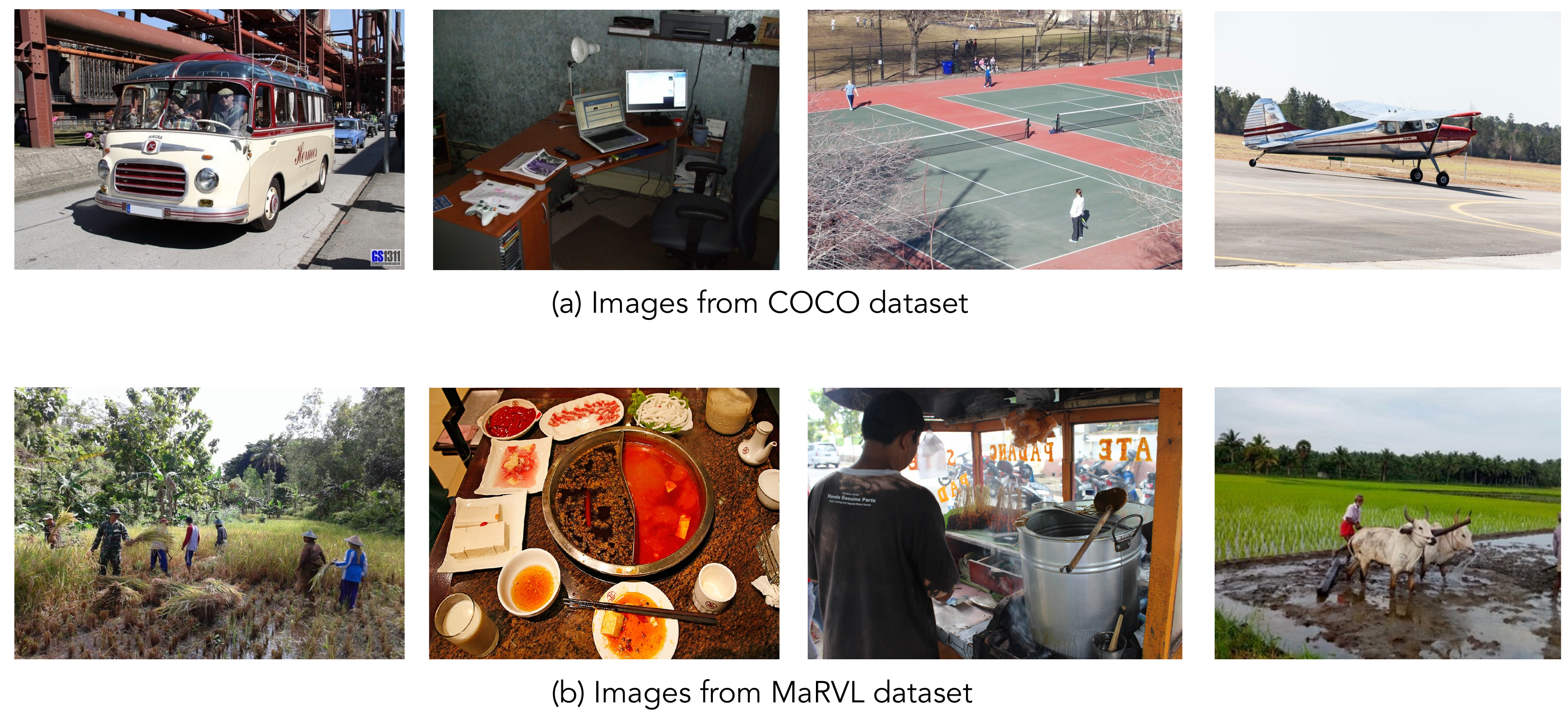}}
%\vspace{-0.5cm}
\caption{Images random sampled from the COCO dataset (top) and the MaRVL dataset (bottom).}
\label{Fig:model}
\end{center}
%\vspace{-0.2cm}
\end{figure*}

\benchmark is a multi-task multi-dimension vision-language benchmark with the goal of providing an accessible platform and a standard practice for the evaluation of VLP models. In this section, we first describe the tasks and datasets included in the \benchmark benchmark. We then introduce two additional dimensions for evaluating new models: generalization ability and efficiency performance trade-off. We describe our private out-of-distribution (OOD) test set that is annotated on images from wild distribution. Finally, we introduce the idea and protocol for evaluating the efficiency-performance trade-off of VLP models.

\subsection{Tasks}

\benchmark consists of five fundamental tasks requiring vision-language understanding and reasoning. We give an overview of all tasks and the corresponding datasets in Table \ref{tab:tasks}, and describe the details as follows:

\paragraph{Image-Text Retrieval} There are two subtasks: text retrieval (TR), where images are queries and texts are targets, and image retrieval (IR), where texts are queries and images are targets. The performance for both subtasks is measured by R$@$K (recall with top k predictions). This task requires VLP models to align the semantic space of vision and language modalities so that two views of a scene are represented similarly in the vector space. 
Previous studies generally evaluate on MSCOCO~\citep{DBLP:conf/eccv/LinMBHPRDZ14} and Flicker30K~\citep{DBLP:journals/ijcv/PlummerWCCHL17}. We opt to MSCOCO because the performance of recent VLP models is saturating on Flicker30K (e.g., ALBEF~\citep{li2021align} achieves 100\% R$@$10 for TR and 98.9\% R$@$10 for IR). 

\paragraph{Visual Reasoning} The natural language visual reasoning task is a binary classification task that takes a pair of images and a natural language statement as input and judges if the statement is true for the image pair. The task requires understanding complex compositional language in visual context. Multiple datasets including NLVR~\citep{DBLP:conf/acl/SuhrLYA17} and CLEVR~\citep{DBLP:conf/cvpr/JohnsonHMFZG17} are collected to test VL models' reasoning ability. We follow the common practice of previous studies on VLP and use the NLVR2 dataset~\citep{DBLP:conf/acl/SuhrZZZBA19} for the visual reasoning task because it uses real photographs in contrast to the other datasets which use synthetic and unrealistic images. Per-example accuracy is used for evaluation.

\paragraph{Visual Grounding} It is another fundamental VL task that aims to locate an object instance from an image with a natural language referring expression. As introduced by~\citet{DBLP:conf/eccv/YuPYBB16}, there exist three variants of the visual grounding task: RefCOCO, RefCOCO+, and RefCOCOg. We opt to the RefCOCO+ subtask, which adds the constraint that no location words are included in the referring expression, because it is of medium difficulty and used in several existing VLP studies~\citep{li2021align,DBLP:journals/corr/abs-2111-08276}. Per-example accuracy is used for evaluation.
%RefCOCO and RefCOCO+ are collected using the ReferItGame, a two-player game where the first player is shown an image with a segmented target object and asked to write a natural language expression referring to the target object and the second player is shown only the image and the referring expression and asked to click on the corresponding object. The difference is that in RefCOCO+, players are not allowed to use location words like ``left''. RefCOCOg is instead collected in a non-interactive setting where annotators are asked to write 

\paragraph{Visual Question Answering} It requires the model to take as input an image and a free-form, open-ended, natural language question about the image and produces or selects a natural language answer as the output. Compared to the image-text retrieval tasks and the binary reasoning task, VQA requires a more detailed understanding of the image and complex reasoning~\citep{DBLP:conf/iccv/AntolALMBZP15}. We select the VQA v2.0 dataset~\citep{DBLP:journals/ijcv/GoyalKASBP19} which balanced the popular VQA dataset~\citep{DBLP:conf/iccv/AntolALMBZP15} so that the language bias is reduced and the understanding of visual clues is more important. Following the VQA 2.0 dataset, we use the publicly released VQA evaluation script in the \benchmark benchmark.

\paragraph{Image Captioning} It is a long standing and challenging problem in vision and language research. It requires a model to take an image as input and generate a natural language sentence (i.e., caption) describing the image. We follow the practice of most previous work and use the COCO Caption dataset~\citep{DBLP:journals/corr/ChenFLVGDZ15} for training and evaluation. We use the widely used BLUE~\citep{papineni2002bleu} and CIDEr~\citep{DBLP:conf/cvpr/VedantamZP15} score for evaluation.

\subsection{OOD Test Set}

We argue that the most important problem in the current evaluation protocol of vision-language models is the ``in-distribution bias'' of the datasets used for downstream tasks. As shown in Table \ref{tab:tasks}, the prevalent datasets for 3 out of 4 representative VL tasks are annotated with images collected from the COCO dataset. The only exception is NLVR2, which also relies on synsets from ImageNet to form queries for the Google Image search engine. On the other hand, almost all VLP models are pre-trained using COCO image-text pairs and many of them initialize the vision encoder with ImageNet pre-training. As such, VLP models have already seen the images (or similar images) in the downstream datasets and their corresponding captions after pre-training. Therefore, fine-tuning on these VL datasets only measures the transferability of VLP models in a special ``in-distribution'' scenario where at least a part of data distribution (i.e., image distribution) remains the same when transferring to a different label distribution. This will likely result in an overestimation of the true generalization ability of VLP models.

To address this problem, we collect a suite of test sets annotated with images from ``in-the-wild'' distribution~\citep{DBLP:conf/icml/KohSMXZBHYPGLDS21} for each of the four representative downstream VL tasks. It is very challenging to find a suitable image source that is sufficiently different from the COCO and ImageNet image distribution while suitable for serving as examples for the vision-language tasks.

Fortunately, \citet{DBLP:conf/emnlp/0001BPRCE21} recently release the MaRVL dataset which provides images from a diverse distribution across cultures and geographical location. In MaRVL, the languages and language-specific concepts are carefully selected by~\citet{DBLP:conf/emnlp/0001BPRCE21} together with members of a community of
native speakers to better mitigate the limitation of ImageNet/COCO concepts and better represent a true worldwide distribution. As such, the MaRVL images significantly differ from the ImageNet/COCO hierarchy which is created in English language culture. We present four randomly selected images from the COCO dataset and MaRVL dataset in Figure \ref{Fig:model}. We can see that images in the COCO dataset clearly differ from those in the MaRVL dataset in terms of both concepts and styles. Therefore, we believe that MaRVL images can serve as a good testbed for measuring the true generalization ability of VLP models.

While MaRVL provides a great source of images for measuring the true transferability of VLP models, it is only annotated on the visual reasoning task with questions written in native languages instead of English. We extend the annotations to support all VL tasks included in \benchmark by crowdsourcing. We hire crowdsource workers that are of good English proficiency from a crowdsourcing platform. The workers are asked to accomplish four annotation tasks to provide test sets for the four downstream VL tasks. We describe them as follows:

\noindent \textbf{Caption Annotation} The workers are asked to write a sentence describing the scene following the instructions from the original COCO caption dataset. To be specific, the sentence is required to have at least 8 words, describe all and only important parts of the image, and does not start with ``there is''. The images and annotated captions are used for the image-text retrieval task.

\noindent \textbf{Statement Annotation} The natural language statements provided by MaRVL for the visual reasoning task are written in native languages. However, most VLP models are trained with English image-text pairs only. Therefore, we need to translate these statements into English. In the original work of~\citep{DBLP:conf/emnlp/0001BPRCE21}, the statements are translated to English by neural machine translation in the Google Cloud API. However, we find that the translation quality of google translation API is not satisfactory for the low-resource native languages in MaRVL (e.g., Indonesian, Swahili, Tamil). To ensure the performance gap between the original in-distribution test set and our OOD test set only comes from the distribution shift instead of translation quality issues, we ask annotators to translate the original statements into English. The workers are given the input image pair and a Chinese translation of the statement generated by Google Translation API, which may of low quality but mostly faithful. They are asked to write an English version of the statement that is both fluent faithful to the original statement.

\noindent \textbf{VQA Annotation} In VQA annotation, the workers are asked to write an open-ended natural language question and the corresponding answer given an image. Inspired by the practice of~\citet{DBLP:journals/ijcv/GoyalKASBP19} for reducing the language bias of VQA datasets and making visual clues matters more, we show a pair of images\footnote{We re-use the image pairs in the MaRVL dataset} of the same concept to the workers and ask them to write a question for which the answers for the two images are different. To ensure the performance gap between the original test set and our annotated OOD test set comes from different image distribution instead of answer distribution, we ask workers to write questions following the question type distribution in the original VQA test set. Also, most VLP models consider the VQA task as a classification problem and only produce labels from a pre-defined answer list consisting of 3129 most frequent answers. Therefore, we ask the workers to verify that the answer for the questions are in the answer list so that all examples in the OOD test set is answerable by the model fine-tuned on the VQA 2.0 train set.

\noindent \textbf{Visual Grounding Annotation} In this task, two groups of crowdsource workers are hired. Workers from the first group are asked to decide if there exist multiple instances of an object and (if so) draw a bounding box of a randomly selected instance. Workers from the other group are asked to write a referring expression for the selected instance following the RefCOCO+ instructions.  

For all annotation tasks, a training session is conducted before annotation so that the workers are carefully trained to ensure they understand the annotation protocol as well as the vision-language tasks the annotations will be used for. We provide the English translation of concept names in native languages for each example to help workers better understand the scene and write annotations. Workers are allowed to skip an instance if they find it too hard or not appropriate for the task. The number of instances in the OOD test set for each VL task is presented in Table \ref{tab:tasks}.

\subsection{Efficiency-Performance Trade-off}

Another limitation of the current practice for VLP evaluation is that the efficiency-performance trade-off is often neglected. Similarly to the trend in the field of pre-training for NLP, VLP models are growing larger and larger in size, which makes it hard, if not impossible, for researchers from most institutes to reach the state-of-the-art in terms of absolute performance. Consequently, more works will instead pursue improvements on other dimensions such as the efficiency-performance trade-off. Therefore, we include the efficiency-performance trade-off as another factor to consider in the \benchmark benchmark.

Measuring and comparing the efficiency-performance trade-off of various VLP models is not straightforward. In \benchmark, we follow the practice of ELUE~\citep{DBLP:journals/corr/abs-2110-07038} and measure a Pareto front of the efficiency-performance trade-off of existing VLP models. In this way, we can easily determine if a new model achieves a Pareto SOTA by seeing if it appears outside of the current Pareto front, i.e., there's no existing model that outperforms the new model on both the performance and efficiency dimensions.

The performance can be easily measured by averaging task-specific metrics across a pre-defined set of tasks. For efficiency, there are three common choices to measure the efficiency of a pre-trained model in the field of NLP: number of parameters~\citep{DBLP:conf/iclr/LanCGGSS20,DBLP:journals/corr/abs-1910-01108}, FLOPs~\citep{DBLP:conf/emnlp/JiaoYSJCL0L20}, and actual inference time~\citep{DBLP:conf/acl/SchwartzSSDS20,DBLP:conf/nips/ZhouXGM0W20}. In \benchmark, we opt to the actual inference time following the Long Range Arena benchmark~\citep{DBLP:conf/iclr/Tay0ASBPRYRM21} for efficient transformers because the latency is more critical for most application scenarios while models with the same number of parameters or FLOPs can lead to very different latency because of their difference in network architectures and parametrizations designs (e.g., going deeper or wider).

In sum, \benchmark is different from previous popular benchmarks because it takes not only the absolute performance but also the generalization ability and the efficiency-performance trade-off into consideration. We believe this will enable the vision and language community to better tell whether we are making genuine progress or overfitting to entrenched datasets, and guide new research in the field to focus on building more generalizable and efficient VLP models instead of overly focusing on the absolute numbers.

%% file: sections/4_experiments.tex
\section{Experiments}

\begin{table*}[t!]
\centering
\resizebox{1.0\textwidth}{!}{
    
\begin{tabular}{l l  cc c c c  c c c }
    % \multicolumn{2}{c|}{\multirow{2}{*}{Tasks}} &\multirow{2}{*}{\footnotesize{SOTA}} & \multirow{2}{*}{\footnotesize{ViLBERT}} & VLBERT & Unicoder & \multirow{2}{*}{\footnotesize{VisualBERT}} &\multirow{2}{*}{LXMERT} & \multicolumn{2}{c}{UNITER}\\
    
\toprule
 \multirow{2}{*}{Tasks} &  \multirow{2}{*}{Metrics}  & ViLBERT & LXMERT & UNITER & VL-T5 & ALBEF & X-VLM & X-VLM  & METER \\
 &  & 274M/3.3M & 240M/180K & 300M/4M & 224M/180K & 210M/14M & 216M/4M & 216M/16M & 352M/4M \\
 % & & ViLBERT & LXMERT & 4M & VinVL & VL-T5 & ALBEF & X-VLM  & METER \\
 \midrule
 \multirow{6}{*}{TR} & R@1 & - & - & 65.68 & -  & 77.60 & 80.40 & 81.20 & 76.16 \\
  & R@5 & - & - & 88.56 & -  & 94.30 & 95.50 & 95.60 & 93.16 \\
  & R@10 & - & - & 93.76 & -  & 97.20 & 98.20 & 98.20 & 96.82 \\
   %\cline{2-10}
  & R@1-OOD & - & - & 36.32 & - & 64.11 & 64.11 & 67.39 & 62.11 \\
  & R@5-OOD & - & - & 63.81 & -  & 88.40 & 88.62 & 89.95 & 87.47 \\
  & R@10-OOD & - & - & 75.13 & - & 93.96 & 94.46 & 95.23 & 92.23 \\
\midrule
 \multirow{6}{*}{IR} & R@1 & 42.51*  & - & 52.93 & - & 60.70 & 63.10 & 63.40 & 57.08  \\
& R@5 & 71.18* & - & 79.93 & - & 84.30 & 85.70 & 85.80 & 82.66 \\  
& R@10 & 80.76* & - & 87.95 & -  & 90.50 & 91.60 & 91.50 & 90.07 \\
   %\cline{2-10}
& R@1-OOD & 27.51 & - & 29.73 & -  & 50.08 & 51.53 & 53.88 & 45.66 \\
& R@5-OOD & 53.07 & - & 56.00 & - & 77.72 & 79.00 & 81.25 & 75.12 \\
& R@10-OOD & 63.87 & - & 66.93 & - & 85.68 & 86.71 & 88.38 & 85.28 \\
\midrule
 \multirow{2}{*}{\small{NLVR$^2$}} & Dev & 77.40* & 74.90 & 79.12  & 73.11* & 82.55 & 84.16 & 84.41 & 82.33 \\
 & Test-P & 78.03* & 76.20 & 79.98  & 73.6 & 83.14 & 84.21 & 84.76 & 83.05 \\
  %\cline{2-10}
 & Test-OOD & 66.53 & 65.24 & 66.60  & 73.84 & 73.17 & 74.16 & 73.02 & 73.47 \\
\midrule
\multirow{3}{*}{\small{VQA}} & Test-dev & 72.70 & 69.90 & 73.82 & 69.80* & 75.84 & 78.07 & 78.22 & 77.68 \\
 & Test-std & - &  72.50 & 74.02  & 70.30 & 76.04 & 78.09 & 78.37 & 77.64 \\
   %\cline{2-10}
 & Test-OOD & 48.38 & 46.18 & 47.79 & 46.31 & 51.52 & 51.55 & 52.57 & 53.76 \\
\midrule
    %  \multirow{4}{*}{\specialcelll{Image\\ Captioning}} & CIDEr & - & - & - &-& 116.5 &  - & - & - \\
    %  & CLIPScore & - & - & - &-& -&  - & - & - \\
    %   %\cline{2-10}
    %   & CIDEr-OOD & - & - & - & - & 36.26 & - & - & - \\
    %   \midrule
    %  & CLIPScore-OOD & - & - & - &-& -&  - & - & - \\
 \multirow{4}{*}{\specialcelll{Visual \\ Grounding}} 
  & Val$^d$ & 74.49* & - & 75.31 & 67.72* & $58.46\star$ & 80.17 & 84.51 & - \\
 & Test-A$^d$ & 79.18* & - & 81.30 & 75.23* & $65.89\star$ & 86.36 & 89.00 & - \\
 & Test-B$^d$ & 66.70* & - & 65.58  & 57.86* & $46.25\star$ & 71.00 & 76.91 & -\\
  %\cline{2-10}
 & Test-OOD$^d$ & 54.91 & - & 36.86  & 27.89 & $24.30\star$ & 55.00 & 59.32 & - \\
 
 %%%%% zengyan: 我这里 打星，是 albef 提出的一种 弱监督的 refcoco 设定！
 
 \bottomrule
\end{tabular}
}
\caption{Results of representative VLP models on \benchmark. X-OOD denotes the results on our private OOD test sets. IR denotes Image Retrieval and TR denotes Text Retrieval. For each compared model, we report the number of parameters and the number of images used for pre-training under the model name. Results with $*$ are our reproduced numbers which are not reported in the original paper. Results with $\star$ are in the weakly supervised setting.}
\label{tab:results}
\end{table*}

\subsection{Training and Evaluation Setup}

We conduct experiments to benchmark the generalization ability and efficiency-performance trade-off of representative VLP models. For each model, we fine-tune the released pre-trained checkpoint on the \benchmark tasks with the hyperparameters provided in the paper. We only consider tasks for which the original paper reported results. This is done for two reasons. First, fine-tuning on different tasks involves various design choices for which we are not able to optimize. Second, some of the models are not suitable for specific kind of tasks (e.g., encoder-decoder models are not suitable for image-text retrieval). We successfully reproduced the performance reported in the original work for all compared models and all tasks. 

After fine-tuning, we evaluate the performance of fine-tuned models on the corresponding OOD test sets in the zero-shot fashion. In addition to the absolute performance, we also record the actual inference time of different models in a controlled setting where the hardware environment is fixed for all models. 
%We report the examples per second as the efficiency metric.

\subsection{Baselines}

We evaluate a number of representative VLP models that effectively learn cross-modal representations and achieve competitive results on many VL tasks. We briefly describe the evaluated models as follows:

\noindent \textbf{ViLBERT}~\citep{DBLP:conf/nips/LuBPL19,Lu_2020_CVPR} ViLBERT is among the first VLP models. It is a two-stream transformer-based model with image representations obtained from object detectors and models cross-modal interactions via co-Attentional transformer layers. It is pre-trained on the Conceptual Captions dataset~\citep{DBLP:conf/acl/SoricutDSG18} with the masked multi-modal modeling objective and the multi-modal alignment prediction objective.

\noindent \textbf{LXMERT}~\citep{DBLP:conf/emnlp/TanB19} LXMERT is also among the first VLP models. The model architecture is similar to that of ViLBERT with a few slightly different designs. LXMERT does not include the multi-modal alignment prediction objective but added the RoI feature regression objective. LXMERT is pre-trained on captions of COCO and VG datasets and also the train/dev set of VQA 2.0, GQA~\citep{DBLP:conf/cvpr/HudsonM19}, and VG-QA~\citep{DBLP:conf/cvpr/ZhuGBF16}.

\noindent \textbf{UNITER}~\citep{chen2020uniter} UNITER is a single stream transformer-based model similar to VisualBERT~\citep{DBLP:journals/corr/abs-1908-03557} and Unicoder-VL~\citep{DBLP:conf/aaai/LiDFGJ20}, which concatenates image representations with text representations to form a single sequence and use transformer layers for representation learning. It is pre-trained with the combination of previous objectives with a modification of applying masking to only one modality at one time, and a novel word-region alignment objective. UNITER is pre-trained on the combination of four caption datasets including COCO captions, VG dense captions, Conceptual Captions, and SBU captions~\citep{DBLP:conf/nips/OrdonezKB11}.

%\noindent \textbf{VinVL}~\citep{DBLP:conf/cvpr/ZhangLHY0WCG21} VinVL is based on Oscar~\citep{DBLP:conf/eccv/Li0LZHZWH0WCG20}. Compared to UNITER, VinVL includes the object tags as vision input, introduced a novel 3-way contrastive loss, and developed a new object detection model that can produce better visual features of images. VinVL is pre-trained with the same set of caption datasets as UNITER while also included image tagging datasets with machine-generated captions and human-annotated tags from a subset of OpenImages.

\noindent \textbf{VL-T5}~\citep{DBLP:conf/icml/ChoLTB21} Different from the above encoder-only models, VL-T5 is based on the sequence-to-sequence framework~\citep{DBLP:conf/nips/SutskeverVL14} and transforms the pre-training objectives and downstream tasks into a unified text generation framework following T5~\citep{raffel2019exploring}. VL-T5 is composed of a single stream transformer encoder with visual embeddings obtained from an object detector and an autoregressive transformer decoder. It is pre-trained on the same set of data as in LXMERT with a combination of multimodal LM, VQA, Image-text matching, Visual Grounding, and captioning as pre-training tasks.

\noindent \textbf{ALBEF}~\citep{li2021align} ALBEF is a two stream transformer-based encoder model. In contrast to the previous models, ALBEF does not require bounding box annotations, thus alleviating the use of an object detector. Instead, it obtains visual embeddings with a visual transformer~\citep{DBLP:conf/iclr/DosovitskiyB0WZ21}. It is pre-trained with the multi-modal language modeling objective, the image-text matching objective, and a new image-text contrastive loss inspired by MoCo~\citep{DBLP:conf/cvpr/He0WXG20}. ALBEF also proposed a momentum distillation method to improve pre-training on noisy image-text pairs. It is pre-trained on COCO, VG, SBU, Conceptual Captions, and an additional dataset, named CC-12M~\cite{changpinyo2021conceptual}.

\noindent \textbf{X-VLM}~\citep{DBLP:journals/corr/abs-2111-08276} X-VLM proposes to learn multi-grained vision language alignments in pre-training. Unlike methods relying on object-centric features, e.g. UNITER and VL-T5, X-VLM relieves the need of object detection and is able to directly leverage the learned multi-grained alignments in downstream tasks. X-VLM is optimized by: 1) locating visual concepts in the image given associated texts by a combination of box regression loss and intersection over union loss; 2) in the meantime aligning the texts with the visual concepts by a contrastive loss, a matching loss, and a masked language modeling loss, where the alignments are in multi-granularity. The pre-training dataset of X-VLM is the same as UNITER.

\noindent \textbf{METER}~\citep{DBLP:journals/corr/abs-2111-02387} 
METER investigates on the transformer-based model designs extensively and proposes a well-designed VLP architecture, \textsc{Meter-CLIP-ViT}.
It applied CLIP-ViT \cite{DBLP:conf/icml/RadfordKHRGASAM21} as the image encoder and RoBERTa \cite{liu2019roberta} as the text encoder.
On top of each encoder, there is a 6-layer transformer to model the cross-modal interaction based on a cross-attention block.
The pre-training objectives are masked language modeling and image-text matching only. 
The pre-training datasets are the same as UNITER.

    % \begin{table*}[]
    % \centering
    % \caption{Main Results}
    % \resizebox{\textwidth}{!}{%
    % \begin{tabular}{l c c c c l l l}
    % \toprule
    % Models & \multicolumn{2}{c}{Image-Text Retrieval} & \multicolumn{2}{c}{Image Captioning} & Visual Reasoning & Visual Grounding & Visual Question Answering  \\
    
    %  & IR & TR & CIDEr & CLIPScore & & & \\  \midrule
    
    % \bottomrule
    % \end{tabular}%
    % }
    % \label{tab:tasks}
    % \end{table*}

\subsection{Results}

\paragraph{Generalization Gap} We present the main result of \benchmark in Table \ref{tab:results}. We find that there exists a sizable generalization gap for all models between the original in-domain tests and our OOD test sets. The state-of-the-art VLP models that seemingly succeed several VL tasks including visual reasoning and VQA (with ~80\% accuracy) still struggle when generalizing to examples from a more diverse distribution. Specifically, the R@1 for the best performing model on the image-text retrieval task drops from 80.4 to 64.1 for text retrieval and from 63.1 to 51.5 for image retrieval. Similarly, for VL reasoning tasks, the best performance drops from 84.2 to 74.2 for NLVR and from 78.1 to 51.7 for VQA. This suggests that the performance of current VLP models is probably overestimated because of the aforementioned in-distribution bias. 

\paragraph{Generalization vs. Performance} In addition, we find that compared to reporting the results in a single dimension (i.e., absolute performance), including the OOD generalization performance into account sometimes leads to complementary and even controversial conclusions. For example, UNITER seems to significantly outperform ViLBERT on both visual reasoning and VQA tasks with an absolute improvement of 1.9 and 1.1 points on the original test sets. However, ViLBERT performs comparably with UNITER on the OOD test set of NLVR2 and even outperforms UNITER by 0.7 points on the OOD test set of VQA. Moreover, VL-T5 performs 9.5 points worse than ALBEF on the in-domain test set of NLVR2 while outperforming ALBEF by 0.7 points on the OOD test set. Similar trends can also be observed when comparing ALBEF, X-VLM, and METER on VQA, and comparing ViLBERT and UNITER on Visual Grounding. These observations demonstrate the necessity of evaluating on the OOD test sets when comparing different models. 

%Moreover, we find that the performance of compared models on the OOD test set of NLVR2 is substantially better (66.5 v.s 63.5 for ViLBERT and 65.2 v.s 62.2 for LXMERT) than that reported by~\citet{DBLP:conf/emnlp/0001BPRCE21}. This confirms that our human translated statements are of better quality compared to the machine translated ones in the original MaRVL dataset. 

\begin{figure}[t]
\begin{center}
\includegraphics[scale=0.5, trim= 0 0 0 0,clip]{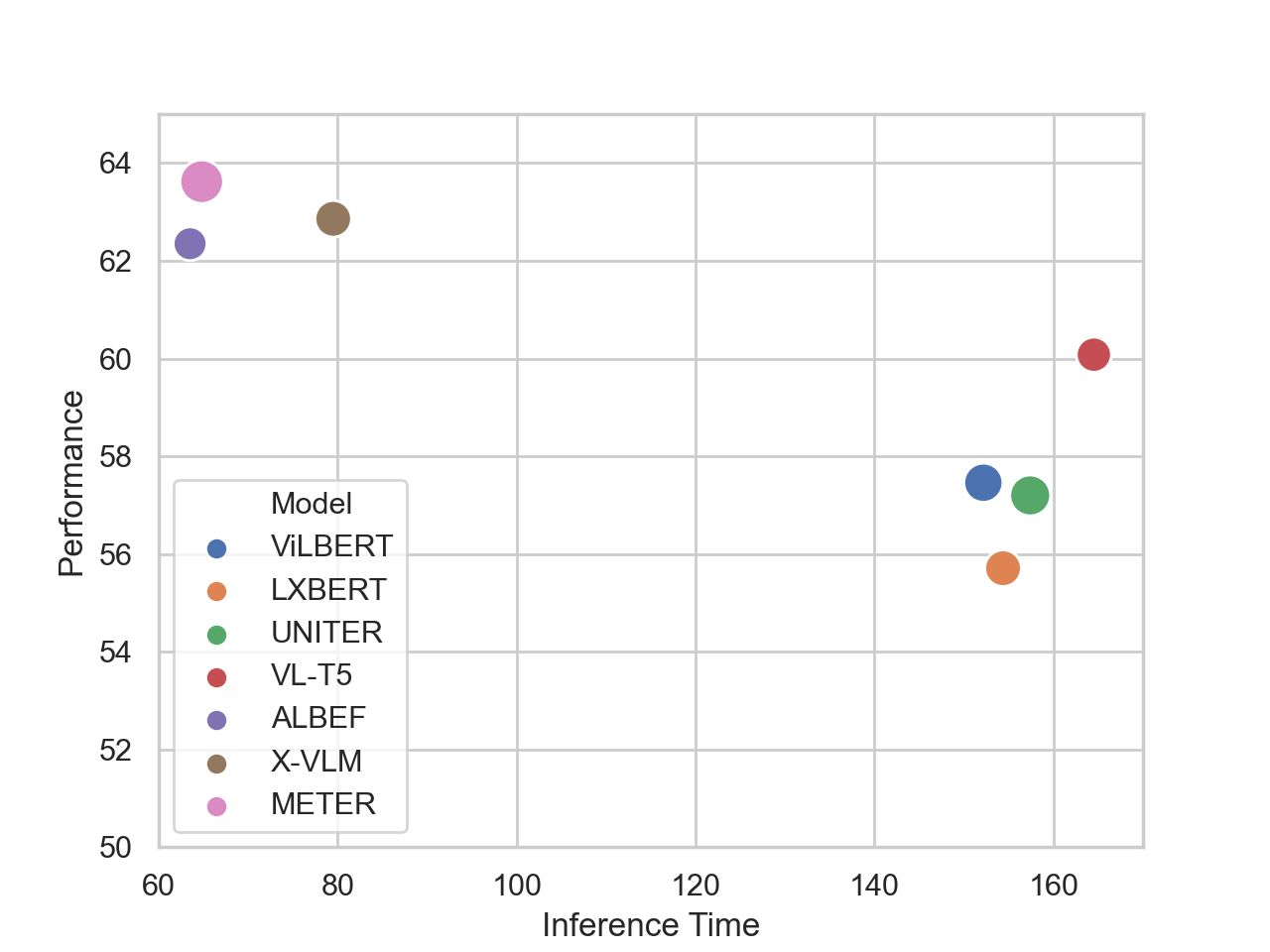} 
\caption{Pareto front of the efficiency-performance trade-off across seven models.}
\label{fig: E-P-tradeoff}
\end{center}
\end{figure}

\paragraph{Efficiency-Performance Trade-off}

We then present the Pareto front in terms of the efficiency-performance trade-off of VLP models in Figure \ref{fig: E-P-tradeoff}. The performance of considered VLP models is measured by the mean accuracy on NLVR2 and VQA test sets. We measure the efficiency of VLP models by their average inference time on these test sets. Note that for models requiring object detection to extract features from the raw images, we include the object detection time in the total inference time. We fix the hardware environment to 1 Nvidia Tesla V100 GPU and the batch size to 1 to simulate real application scenarios\footnote{The actual inference time of different models depends on hardware. We provide the code for measuring actual inference time to facilitate researchers compare efficiency of different compared models in their own envirnment. We also provide the actual inference time of popular vision-language models in our setting in the Appendix of our paper for reference.}. We can see that models with similar performance may have completely different positions in the Pareto front of efficiency-performance trade-off. For example, VL-T5 performs only marginally worse than ALBEF, but requires an average inference time 2.4 times longer than ALBEF. Therefore, ALBEF should be considered to significantly outperform VL-T5 in the dimension of efficiency-performance trade-off.
This further demonstrates the necessity of a multi-dimension benchmark like \benchmark for more throughout comparisons.

\begin{table}[]
    \centering
    \resizebox{1.0\columnwidth}{!}{
    \begin{tabular}{l|cccc}
    \toprule
         Models &  BLEU4 & CIDER & BLEU4-OOD & CIDER-OOD \\
    \midrule
       \bf VL-T5  & 34.2 & 113.7 & 10.9 & 36.6 \\
       \bf X-VLM & 39.9 & 134.0 & 17.4 & 67.4 \\
       \bf OFA & 41.7 & 140.7 & 18.7 & 71.2  \\ 
       \bottomrule
    \end{tabular}}
    \caption{Results on Image Captioning.}
    \label{tab:caption}
\end{table}

\paragraph{Results on Image Captioning}

Most aforementioned models are encoder-only models and thus not suitable for image captioning. Therefore, we also include OFA~\citep{wang2022unifying}, the state-of-the-art pre-traiend VLM for image captioning. We present the results on image captioning in Table \ref{tab:caption}. We can see that the performance of all evaluated model drops significantly, which is in line with the previous results on vision-language understanding tasks. Moreover, we find that VL-T5, which is based on object detection for visual representation, suffers from a larger performance drop. We conjecture this is because object tags and embeddings from a pre-trained object detector plays an important role in its success on the COCO caption dataset, and the object detector may fail to generalize well on more diverse concepts in our OOD test set.

\section{Analysis}

We conduct a series of analyses investigating the relationship between different evaluation dimensions in \benchmark and the influence of different design choices for VLP models on these dimensions.

\paragraph{Performance-Generalization Correlation} We calculate the Pearson correlation coefficient $\rho$ of the models' performance on in-domain test sets and that on our OOD test sets. We obtain a relatively high correlation ($\rho = 0.75$). This suggests that models perform well on in-domain test sets also tend to succeed on the OOD test set. However, the correlation is not very high, which indicates that there may be some differences between the trend of in-domain performance and OOD performance. This also supports the need of a multi-dimensional benchmark like \benchmark.

\paragraph{Object Detection vs. Vision Transformer} We investigate the impact of vision feature sources on \benchmark. We find that models with different vision feature sources such as object detection features or directly using vision transformers only slightly differ in terms of absolute performance. However, from the Pareto front of efficiency-performance trade-off, we can clearly find that models with vision transformers as image encoders are more practical in terms of efficiency-performance trade-off.

%% file: sections/5_conclusion.tex
\section{Conclusion and Discussion}
We introduce \benchmark, a multi-task multi-dimension benchmark for the evaluation of vision-language pre-trained models. \benchmark is a first step towards evaluating a model not only by its absolute performance but also on other useful dimensions including the generalization ability of the model and its efficiency. We include four representative vision-language tasks/datasets in \benchmark. For each task, we crowdsource an OOD test set annotated with images from a diverse distribution to measure the generalization gap. We benchmark the performance, generalization ability and efficiency-performance trade-off of 7 representative VLP models to facilitate future research. In sum, \benchmark aims to promote vision-and-language research that does not solely focus on absolute performance but also takes other important factors like generalization and efficiency into account.

In addition, we would also like to highlight that while the original intended use of the collected OOD data in \benchmark is to evaluate vision-language models via direct OOD generalization (i.e., fine-tune on original datasets then directly test on OOD test sets), it is also possible to fine-tune, or few-shot fine-tune on a subset of the provided OOD data for other research settings such as transfer learning and domain adaptation.